%% file: athena.tex
\pdfoutput=1

\documentclass[11pt]{article}

\usepackage{EMNLP2023}

\usepackage{times}
\usepackage{latexsym}

\usepackage[T1]{fontenc}

\usepackage[utf8]{inputenc}

\usepackage{microtype}

\usepackage{inconsolata}

\usepackage{multirow}
\usepackage{graphicx}
\usepackage{booktabs}
\usepackage{enumitem}
\usepackage{amsmath}
\usepackage{amsthm}
\usepackage{amsfonts}
\usepackage[bb=dsserif]{mathalpha}
\usepackage{algorithm}
\usepackage{algpseudocode}
\usepackage{tikz}
\usepackage{pgfplots}
\usepackage{pgfplotstable}
\pgfplotsset{compat=1.9}
\usepackage{amssymb}

\DeclareMathOperator*{\argmax}{arg\,max}
\DeclareMathOperator*{\attn}{\mathrm{A}}
\DeclareMathOperator*{\sattn}{\attn_{\mathrm{self}}}
\DeclareMathOperator*{\merge}{\mathrm{M}}
\DeclareMathOperator*{\tran}{\mathrm{T}}

\newcommand{\layernorm}{\ell}
\newcommand{\ff}{\operatorname{FF}}
\newcommand{\R}{\mathbb{R}}
\newcommand{\where}{\mathrm{where}}
\newcommand{\st}{\text{s.t.}}
\newcommand{\op}{\operatorname{op}}

\newcommand{\PV}{\mathrm{P}}
\newcommand{\GV}{\mathrm{G}}
\newcommand{\infer}{\operatorname{infer}}
\newcommand{\answer}{\operatorname{answer}}
\newcommand{\Eq}{\mathcal{E}}
\newcommand{\1}{\mathbb{1}}

\title{ATHENA: Mathematical Reasoning with Thought Expansion}

\author{
JB. Kim\textsuperscript{1} \qquad
Hazel H. Kim\textsuperscript{2} \qquad
Joonghyuk Hahn\textsuperscript{1} \qquad
Yo-Sub Han\textsuperscript{1}\\
\textsuperscript{1}Yonsei University \qquad
\textsuperscript{2}Classting AI Research  \\
\texttt{
jb@thejb.net,
hazel.kimh@gmail.com,
\{greghahn,emmous\}@yonsei.ac.kr} 
}

\begin{document}

\maketitle
\begin{abstract}
Solving math word problems depends on how to articulate the problems, the lens through which models view human linguistic expressions.
Real-world settings count on such a method even more due to the diverse practices of the same mathematical operations.
Earlier works constrain available thinking processes by limited prediction strategies without considering their significance in acquiring mathematical knowledge.
We introduce \textbf{A}ttention-based \textbf{TH}ought \textbf{E}xpansion \textbf{N}etwork \textbf{A}rchitecture (\textbf{ATHENA}) to tackle the challenges of real-world practices by mimicking human thought expansion mechanisms in the form of neural network propagation.
A thought expansion recurrently generates the candidates carrying the thoughts of possible math expressions driven from the previous step and yields reasonable thoughts by selecting the valid pathways to the goal.
Our experiments show that ATHENA achieves a new state-of-the-art stage toward the ideal model that is compelling in variant questions even when the informativeness in training examples is restricted.\footnote{The source code is available at \url{https://github.com/the-jb/athena-math}.}
\end{abstract}

\section{Introduction}

\begin{figure}[t]
    \centering
    \includegraphics[width=\columnwidth]{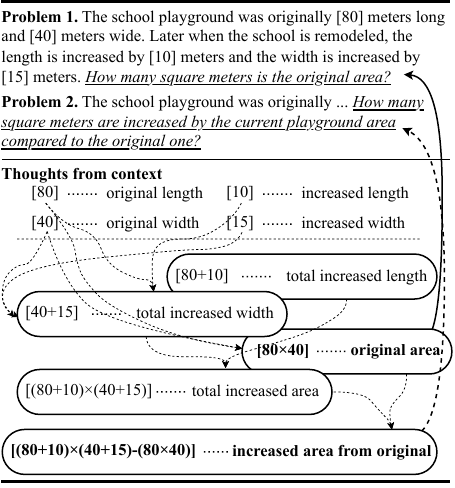}
    \vspace{-1.5em}
    \caption{Visualization of thoughts constructed for solving two problem samples with the same context description from the UnbiasedMWP dataset, one of our benchmarks.}
    \label{fig:thoughts}
\end{figure}

\input{figures/table-dr-athena.tex}

Math word problem (MWP) solving is one of the fundamental reasoning tasks of answering a mathematical question by understanding a complex, intricate system of human lexical expressions.
Models' ability to solve a problem depends on a method that articulates the problem, the lens through which they view human lexical expressions.
Ideal MWP models understand the diverse applications of the same mathematical operations in real-world situations, which require lexically sophisticated.
For example, ``$\times$'' can count all elements equally divided in multiple boxes, but also calculate area from length and width, or tax fee from the tax rate and income.

It is significant how we estimate if the model has learned mathematical reasoning to qualify for the ideal model.
We state that the ideal models that learn mathematical problems must be able to solve unseen problems if they are applications of mathematical operations that models have already seen, or soundly solve problems even when given examples to learn are restricted.

Humans learn mathematical knowledge by formulating and understanding underlying principles from seen cases rather than just recognizing the common lexical patterns.
Models currently face two challenges to reach human-level mathematical understanding: \textit{conceptual knowledge} to understand the practices of mathematical principles, and \textit{procedural knowledge} to deductively derive the answer through the principles, which are indispensable for each other in their development and usage~\citep{rittle1999conceptual, byrnes1991role, canobi2009concept, rittle-etal-2014-developing}.

Prior approaches mostly adopt the transduction-based models such as sequence-to-sequence~\citep{ling-etal-2017-program, wang-etal-2018-translating}, sequence-to-tree~\citep{xie-sun-2019-goal, liu-etal-2019-tree} or graph-to-tree methods~\citep{zhang-etal-2020-graph-tree, li-etal-2020-graph-tree} and concentrate on enhancing problem-level encoding~\citep{shen-jin-2020-solving, zhang-etal-2020-graph-tree, lin-etal-2021-hms, yu-etal-2021-improving}.
These works have limitations in obtaining procedural knowledge due to their prediction strategies that operate in a counter-intuitive order.
For instance, sequence-to-tree approaches determine mathematical operations before the operands in the inference steps.
Recently, \citet{jie-etal-2022-learning} proposed a deductive prediction strategy, but it still entails procedural bias, accepting only one particular reasoning pathway.

Overall, the prior studies show limitations in learning mathematical procedures, which is significant for achieving successful mathematical skills.
As a result, despite their high accuracies on some benchmarks, the current approaches fail to solve variant questions that are simply mutated from already trained examples~\citep{patel-etal-2021-nlp, yang-etal-2022-unbiased}.
As shown in Table~\ref{tab:dr_athena}, we empirically argue that they learn the repeated patterns in given problems rather than the underlying principles formulating the equations.

The cognitive inadequacy of previous models motivates us to propose a new reasoning architecture that maximizes feasible reasoning pathways.
We design ATHENA that reasons with thought expansion inspired by the studies of human reasoning
\citep{johnson2008we, rittle-etal-2014-developing}.
The key idea is to implement two types of thoughts in the expansion process: \textit{thoughts before considering goals} formed by conceptual knowledge and \textit{goal-directed thoughts} yielded by procedural knowledge.

Figure~\ref{fig:thoughts} illustrates an example of thoughts formed via asking different questions (goal) within the same situation (context).
Because it is tricky for models to answer different questions that share a lexically similar problem context, we state that the two different thoughts would lead to the right answer by properly utilizing mathematical knowledge.
We expand the thoughts by applying conceptual knowledge to obtain candidate thinking pathways and procedural knowledge to evaluate the potential answers.
This is how we endow models with mathematical reasoning ability and empirically demonstrate that the model has actually learned the knowledge.

ATHENA puts the aforementioned thinking process explicitly into neural network propagation.
Defining the term \textit{thought} as a representation of each math expression driven from the problem, we shape \textit{candidate thoughts} and the goal-directed thoughts named \textit{reasonable thoughts}.
The model generates candidate thoughts by applying mathematical operations and yields reasonable thoughts by filtering with solidly updated premises until it meets the appropriate answer.
With this recurrent process, we develop a neural model of processing thoughts based on multi-head attention~\citep{vaswani-etal-2017-attention} that effectively carries the subtle feature changes during expansion.

Our experiments show that the proposed approach is strong at predicting mathematical expressions requiring sophisticated comprehension as shown in Table~\ref{tab:dr_athena}.
We observe that ATHENA produces a solid performance when the model needs to deal with previously unseen questions.
ATHENA is also very compelling to solve variant questions once it has learned one question established from the shared context.
From the experimental results, we conclude that ATHENA reaches a new state-of-the-art stage toward the ideal MWP model that we define as the one that can learn mathematical reasoning.

\section{Math Word Problem}

Math word problem (MWP) solving is the task of answering a mathematical question by understanding natural language descriptions.

\subsection{Problem Formulation}

Our task of solving MWPs is defined as follows.
Each example in the MWP dataset $\mathcal{D}$ has a problem sequence~$S$ in natural language as input and an equation~$\Eq$ as expected output.
$\mathcal{D}$ consists of $K$~(problem, equation) tuples, where $K$ is the number of examples:
\begin{equation*}
\mathcal{D}=\{(S_{(i)}, \Eq_{(i)})\}_{i=1, \dots , K} .
\end{equation*}
We use a pre-trained language model~(PLM) to embed $S$.
Let $P=(t_1,t_2,\dots,t_n)$ denote a tokenized sequence of $S$,
where $t_i$ represents each subword token.
The PLM output of $P$ is denoted as $X=(x_1,x_2,\dots,x_n)$, where $x_i$ is an embedding vector of each token~$t_i$.

\subsection{Related Work}

MWP problems have begun with feature engineering via hand-crafted rules or statistical concepts~\citep{bakman-etal-2007-robust, hosseini-etal-2014-learning, mitra-baral-2016-learning}.
Early works have adopted neural network approaches through end-to-end learning strategies such as sequence-to-sequence~\citep{ling-etal-2017-program, wang-etal-2018-translating, li-etal-2019-modeling} or sequence-to-tree~\citep{xie-sun-2019-goal, liu-etal-2019-tree, chiang-chen-2019-semantically, qin-etal-2020-semantically}. The previous approaches often 
use the networks or manipulate the representation with tree or graph templates to generate mathematical equations in a structurally sophisticated manner~\cite{wang-etal-2017-deep, zhang-etal-2020-graph-tree}.

Having developed and become accessible to pre-training and transfer learning, several approaches have promoted their performance with pre-trained language models~\citep{shen-etal-2021-generate-rank, yu-etal-2021-improving, huang-etal-2021-disenqnet, zhang-etal-2020-teacher, liang-etal-2022-mwp}, aiming to enhance the encoder with pre-trained embeddings.
Other approaches have made a key contribution by utilizing additional knowledge such as semantic meaning. Some take advantage of structural information such as hierarchical dependency~\citep{shen-jin-2020-solving,lin-etal-2021-hms,yu-etal-2021-improving}, formula structure~\citep{huang-etal-2020-neural}, graph-edge connection information~\citep{zhang-etal-2020-graph-tree,wu-etal-2021-edge-enhanced, li-etal-2020-graph-tree} and more~\citep{li-etal-2022-seeking, shen-etal-2021-generate-rank}.

All these approaches mostly aim at enhancing problem-level information but the recent studies demonstrate the significance of inference procedures in reasoning tasks.
\citet{wei-etal-2022-chain} show impressive success for large-scale language models in complex reasoning tasks by adopting chain-of-thought prompting.
The reasoning extraction method~\citep{jie-etal-2022-learning} has recently reached decent performance by constructing the deductive order in solving MWP.

\section{ATHENA}
\label{sec:athena}
\begin{figure*}[t]
    \centering
    \includegraphics[width=\textwidth]{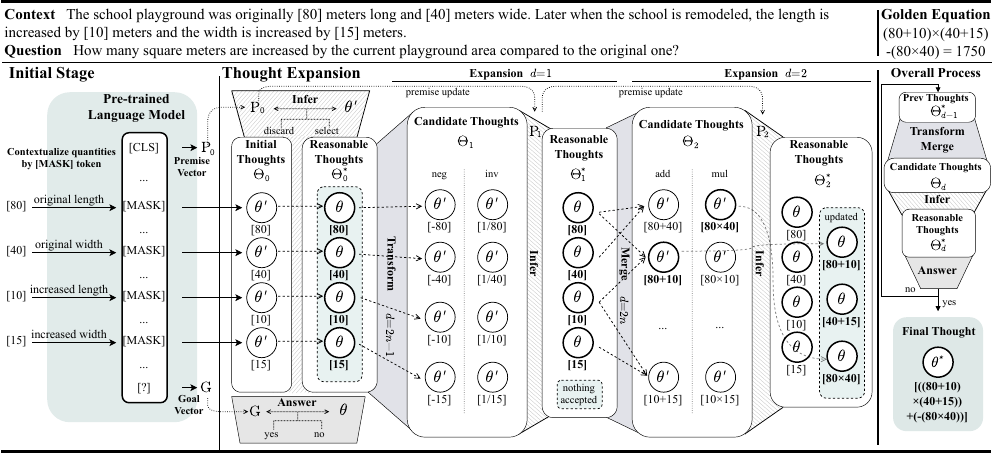}
    \vspace{-1.5em}
    \caption{Overall process of ATHENA. First, extract initial thoughts, an initial reasoning vector, and a goal vector from PLM. Second, expand thoughts by transform $(d=1,3,5,\dots)$ or merge $(d=2,4,6,\dots)$ and generate candidate thoughts. Third, infer the candidate thoughts to obtain new reasonable thoughts. Last, give reasonable thoughts to the next expansion. Repeat until meeting a thought that answers the goal vector.}
    \label{fig:athena_overview}
\end{figure*}

\textbf{A}ttention-based \textbf{TH}ought \textbf{E}xpansion \textbf{N}etwork \textbf{A}rchitecture (ATHENA) is an architecture that expands its thoughts to solve MWP.
Figure~\ref{fig:athena_overview} illustrates an overall process of ATHENA.
ATHENA extracts initial thoughts $\Theta_0$ from PLM and expands them with inferring through premises until it reaches the final thought.
We first clarify what is a 
\textit{thought}---a foundational ingredient of our model---and explain the premise and goal vectors that measure the thoughts.

\paragraph{Thought.}
A thought is an embedding of a possible math expression derived from quantities in a problem representing the contextual meaning of the expression.
Let $\theta$ denote a thought with hidden size $H$ corresponding to an expression $\Eq(\theta)$.
A goal of the model is to find a thought~$\theta^*$ that satisfies the ground-truth expression $\Eq^*$:
\begin{equation*}
\Eq(\theta^*) \equiv \Eq^*.
\end{equation*}

\paragraph{Premise Vector.}
A premise vector represents previously inferred thoughts to evaluate and filter candidate thoughts in each depth.
Let $\PV_d$ denote a premise vector for depth $d$.
We set an initial premise vector $\PV_0$ with the [CLS] token from the problem descriptions.

\paragraph{Goal Vector.}
A goal vector plays a role as ground-truth measurement to evaluate if a thought is an appropriate answer to the question.
We set a goal vector $\GV$ with a tokenized embedding of the punctuation mark in the question description.

\subsection{Initial Thought}
\label{sub:initial_stage}

An initial thought is an embedding that carries each quantity representation illustrated in a context or question description.
We mask quantities with [MASK] token and obtain the embeddings that capture contextual information from the perspective of corresponding quantities.
We denote a set of thoughts in the initial depth by $\Theta_0$:
\begin{equation*}
\Theta_0 = \{x_i \mid x_i \in X, t_i \in P, t_i=\operatorname{[MASK]} \}.
\end{equation*}

Certain quantity representations such as $\pi$ are necessary for generating mathematical expressions despite not being presented in the contexts or questions. We collect them from a training set and randomly initialize their embeddings.
We also put their embeddings to initial thoughts $\Theta_0$.

\subsection{Thought Expansion}
In each depth, thought expansion constructs candidate thoughts $\Theta_d$ and filters them to obtain the \textit{reasonable} thoughts $\Theta_d^*$.
Reasonable thoughts are the waypoint thoughts to reach the final thought.

The two stages in a thought expansion are:
(1) our model generates candidate thoughts $\Theta_d$ from previous thoughts $\Theta_{d-1}^*$ through the operations
and (2) it reasons about the candidates if they are worth to be reasonable thoughts $\Theta_d^*$.
Expansion keeps going until finding one of the reasonable thoughts qualified to be a final thought $\theta^*$.

\subsubsection{Candidate Thought}
Our model generates a set of possible new 
thoughts~$\Theta_d$ from the previous thoughts $\Theta_{d-1}^*$ as the candidates.
A new thought $\theta'$ is a thought of a math expression obtained by combining two previous thoughts $\theta_i,\theta_j\in\Theta_{d-1}^*$ with an arithmetic operation:
\begin{equation*}
\Eq(\theta')=\Eq(\theta_i)\circ\Eq(\theta_j)
\;\where\; \circ\in \{+,-,\times,\div \} .
\end{equation*}

To make a new thought, we introduce two operation layers whose combination can represent the arithmetic operations: merge $\merge$ and transform $\tran$.
These layers aim to maximize the reflection of the characteristics of arithmetic operations rather than the separate layers of individual arithmetic operations.
The definitions of merge and transform are shown below.

\paragraph{Merge.}
Merge layer $\merge$ merges a pair of thoughts $(\theta_i, \theta_j)$ into a new thought $\theta'$ such that $\Eq(\theta')$ applies addition and multiplication to $\Eq(\theta_i)$ and $\Eq(\theta_j)$:
\vspace{-1em}
\begin{multline*}
\merge^{\op}: \theta_i, \theta_j \mapsto \theta' \\
\st\, \Eq(\theta') = \op(\Eq(\theta_i) , \Eq(\theta_j) )
\,\where\, \op \in \{+, \times\} .
\end{multline*}

\paragraph{Transform.}
Transform layer $\tran$ transforms a thought $\theta$ into a new thought $\theta'$ such that
$\Eq(\theta')$ applies inverse operations of addition and multiplication to $\Eq(\theta)$:
\vspace{-1em}
\begin{multline*}
\tran^{\op}: \theta \mapsto \theta' \\
\st\; \Eq(\theta') = \op(\Eq(\theta))
\;\where\; \op \in \{-\cdot, \cdot^{-1}\}.
\end{multline*}

We use Feed-Forward Network (FFN) and multi-head attention inspired by \citet{vaswani-etal-2017-attention} for the implementation of the operation layers.
We use FFN referred to as $\ff$ for transform layer $\tran$.
Using multi-head self-attention $\sattn$ and layer normalization $\layernorm$, we implement merge layer $\merge(\theta_i,\theta_j)$ followed by:
\vspace{-.3em}
\begin{equation*}
\merge(\theta_i,\theta_j) = \ff(\theta_i+\theta_j+\layernorm(\mathbf{1}_2^\top \sattn([\theta_i; \theta_j]))W+b)
\vspace{-.7em}
\end{equation*}
$\hfill \where\; W\in \R^{H\times H}, b\in\R^H .\;\;$
\vspace{.7em}

This implementation satisfies $\merge^{\op}$ to be commutative
for $\op\in\{+,\times\}$:
\begin{multline*}
\merge^{\op}(\theta_i,\theta_j)
=\merge^{\op}(\theta_j,\theta_i) \;\mathrm{and} \\
\Eq(\merge^{\op}(\theta_i,\theta_j))
= \Eq(\merge^{\op}(\theta_j,\theta_i)).
\end{multline*}
We apply transform layer $\tran$ for depth $d=2n-1$
and merge layer $\merge$ for depth $d=2n$ to generate the candidates.
In the case of the beginning depth $d=0$, we use the initial thoughts $\Theta_0$ as the candidates.

\subsubsection{Reasonable Thought}
After obtaining candidate thoughts $\Theta_d$, our model yields reasonable thoughts $\Theta_d^*$ that constitute the final thought $\theta^*$.
In each depth $d$, it selects reasonable thoughts from candidate thoughts through the inference layer $\infer$ with a premise vector $\PV_d$.

\paragraph{Infer.}
The inference layer calculates the correlation score between the premise vector $\PV_d$ and each candidate thought $\theta\in\Theta_d$ using multi-head attention $\attn(Q,K=V)$ and feed-forward network $\ff$ with sigmoid function $\sigma$ to evaluate if a thought is acceptable within the premises:
\begin{multline*}
\infer(\PV_d, \theta) = \sigma(\attn(\ff(\theta),\PV_d)W_r+b_r)\\
\where\; W_r\in\R^{H\times1}, b_r\in\R .
\end{multline*}
A thought $\theta$ is \textit{reasonable} if its correlation score $\infer(\PV_d,\theta)$ exceeds a threshold $t_r=0.5$.
In the next iteration $d+1$, the reasonable thoughts in the current depth $\Theta_d^*$ become the input.

\paragraph{Update Premises.}
A previously obtained consequence can become a premise for the next inference step.
Accordingly, our model updates the premise vector $\PV_d$ with the reasonable thoughts $\Theta_d^*$ obtained in the current depth $d$ to prepare a premise vector for the next step $\PV_{d+1}$.
It gains the updated premise vector by concatenating all reasonable thoughts $\Theta_d^*$ after the multi-head attention $\attn$ using the parameters of the inference layer $\infer$:
\begin{equation*}
\PV_{d+1} = \PV_d \mathbin\Vert \attn(\ff([\Theta_d^*]),\PV_d) .
\end{equation*}

\begin{algorithm}[t]
\renewcommand{\algorithmicrequire}{\textbf{Input:}}
\renewcommand{\algorithmicensure}{\textbf{Output:}}
\small
\caption{\small Thought Expansion Process of ATHENA}\label{alg:athena}
\begin{algorithmic}
\Require $\Theta_0, \PV_0, \GV$
\Ensure $\Eq^*$
\State $d \gets 0$
\State $\Theta^*_0 \gets \{ \theta \mid \theta \in \Theta_0, \infer(\PV_0, \theta) \geq t_r \}$ 
\While{$d\leq D \;\textbf{or}\; \exists\theta\in\Theta^*_d(\answer(G, \theta) > t_{f})$}
\State $\PV_{d+1} \gets \PV_d \mathbin\Vert \attn(\ff([\Theta_d^*]),\PV_d)$
\State $d \gets d+1$
\If{$d=1, 3, 5 \dots$} \State $\Theta_d \gets \bigcup_{\op\in\{-\cdot,\cdot^{-1}\}}\{\tran^{\op} (\theta)\mid\theta \in \Theta^*_{d-1}\}$
\ElsIf{$d=2, 4, 6 \dots$} \State $\Theta_d \gets \bigcup_{\op\in\{+,\times\}}\{\merge^{\op} (\theta_i,\theta_j)\mid \theta_i,\theta_j \in \Theta^*_{d-1} \}$
\EndIf
\State $\Theta^*_d \gets \Theta^*_{d-1} \cup \{\theta\mid\theta\in\Theta_d, \infer(\PV_d,\theta) \geq t_r \}$
\EndWhile
\State $\theta^* \gets \argmax_{\theta \in \Theta^*_d} \answer(G, \theta) $ \\
\Return $\Eq(\theta^*)$
\end{algorithmic}
\end{algorithm}

\subsection{Final Thought}
\label{sec:final}

A final thought $\theta^*$ is the answer to the question.
When the thought expansion process finishes, our model decides the final thought by selecting a thought with the maximum score.
We have two criteria to terminate the iteration;
(1) when the depth reaches the maximum expansion depth $D$;
(2) if there is a thought with the score that exceeds a confidence threshold $t_f$ on iteration.
We calculate the score of each reasonable thought  $\theta\in\Theta_d^*$ using the multi-head attention $\attn$ and feed-forward network $\ff$ with the goal vector $\GV$, activated by sigmoid $\sigma$:
$$
\answer(\GV, \theta) = \sigma(\attn(\ff(\theta),\GV)W_a+b_a),
$$
where $W_a\in\R^{H\times1} , b_a\in\R$.

A thought with the maximum score in the reasonable thoughts becomes a final thought $\theta^*$: 
\begin{equation*}
\theta^*=\argmax_{\theta \in \Theta^*_d}\left( \answer(G, \theta) \right).
\end{equation*}
The model bestows the final thought the fidelity to shape the answer to the goal of the problem.

Algorithm~\ref{alg:athena}
shows the overall process to derive the final answer $\Eq(\theta^*)$ from inputs $\Theta_0, \PV_0, \GV$.

\section{Experiments}

We conduct experiments across a comprehensive range of MWP) solving tasks to show that ATHENA outperforms strong baselines in both full datasets and variant versions of the original datasets while being more interpretable in terms of intermediate steps toward the answers.

\subsection{Experimental Setups}

\paragraph{Baselines.}
We select four representative approaches as the baselines to compare with ATHENA: 
Transformer~\citep{vaswani-etal-2017-attention}\footnote{We follow hyperparameters by \citet{lan-etal-2022-mwptoolkit} for both vanilla transformer and RoBERTa-based transformer.},
a goal-driven tree-structured model (GTS)~\citep{xie-sun-2019-goal},
Graph-to-Tree~\citep{zhang-etal-2020-graph-tree}\footnote{We follow the best hyperparameter settings in \citet{patel-etal-2021-nlp} for both vanilla models and RoBERTa-based models.}
and Deduct\-Reasoner~\citep{jie-etal-2022-learning}.\footnote{We use their hyperparameter setups. We use the MAWPS setup for testing ASDiv-A, and use the Math23k setup for UnbiasedMWP. Since the authors do not provide setups for RoBERTa-large, we optimize the model and report the best score with half batch size and half learning rate from those used in the RoBERTa-base setup.}
Transformer is a sequence-to-sequence approach that uses multi-head attention mechanism while
GTS is a strong baseline of sequence-to-tree model.
Graph-to-Tree is another approach that adds a graph encoder on top of GTS.
We adopt DeductReasoner as an additional baseline that introduces a complex relation extraction method for deductive steps and hence achieves the state-of-the-art performance.

\paragraph{Implementation Details.}
We use RoBERTa-base and RoBERTa-large as our base pre-trained embeddings~\citep{liu-etal-2019-roberta} and Chinese-RoBERTa \citep{chinese-roberta} for Chinese benchmarks to compare our baselines.
We use pre-layer normalization \citep{xiong-etal-2020-layer} for our multi-head attention method to fully leverage a dynamic range of embeddings.
We set $D$ by the maximum value of the reasoning depth of test examples for each dataset.\footnote{We present the values of each dataset in Table~\ref{tab:stop}.}
We set $t_f=0.95$ and train our model by giving ideal accepted prior thoughts $\Theta_{d-1}^*$ and labels of $\infer$ and $\answer$ in each depth to calculate the loss with binary cross entropy over all labels.\footnote{We present detailed training settings and hyperparameters in Appendix~\ref{sec:appendix_training}}
We perform our experiments with Nvidia RTX 3090 GPU.

\paragraph{Dataset.}
We test ATHENA on both standard MWP benchmarks 
and relatively new benchmarks that contain various linguistic expressions in contexts or questions for mathematical reasoning. The standard benchmarks are \textbf{MAWPS}~\citep{koncel-kedziorski-etal-2016-mawps}, \textbf{ASDiv-A}~\citep{miao-etal-2020-diverse}, and \textbf{Math23k}~\citep{wang-etal-2017-deep}. MAWPS is an English corpus collected from the online MWP repository, and
Math23k is a Chinese corpus crawled from online posts. ASDiv-A is an acronym of An arithmetic subset of Academia Sinica Diverse dataset (ASDiv-A), consisting of diverse English lexical patterns.

The relatively new benchmarks either alter the standard benchmarks or vary the grounded expressions from the collected data to evaluate the model performance without bias from learned data. \textbf{SVAMP}~\citep{patel-etal-2021-nlp} varies in the components of one of the standard benchmarks, ASDiv-A to evaluate various contextual expressions on elementary-level arithmetic problems.
\textbf{UnbiasedMWP}~\citep{yang-etal-2022-unbiased} is an online-crawled Chinese corpus that augments the questions from the same context to evaluate models if they are able to generate adequate corresponding mathematical expressions. We split MAWPS, ASDiv-A, Math23k, SVAMP, following \citet{jie-etal-2022-learning} and \citet{patel-etal-2021-nlp}, respectively.

\input{figures/table-main.tex}

\paragraph{One-to-Many Test.}
\label{sec:one_to_many}
In addition to the standard test, we conduct one-to-many variants tests to measure model generalization to many variant questions from one example within the common context.  
We select two datasets SVAMP and Unbiased\-MWP to apply for this test. 
Each example in the dataset has a problem sequence that is composed of context and question descriptions.
Within the groups by context, we split the examples one-to-many.
One randomly selected example per group goes to a training set while the rest examples in the group move to a test set.
We use the examples that do not have other variants within the context group as a validation set.
We name the resorted SVAMP and Unbiased\-MWP using the one-to-many setup as SVAMP(1:N) and Unbiased\-MWP(1:N).
We construct these sets with 5 different random seeds to mitigate training bias and report the average performance.

\subsection{Results}

We repeat our experiments 5 times with different random seeds and report the average answer accuracy with the standard error.
We report results on multiple benchmarks, variants splitting tests, the impact of pre-trained language models depending on their size, and ablation tests.

\paragraph{Overall Performance.}
Table~\ref{tab:main} shows the performance of different methods on 7 benchmarks. ATHENA establishes new state-of-the-art results for overall benchmarks. ATHENA outperforms prior MWP methods on all occasions with one exception of its performance on Math23k when trained on the RoBERTa-base model.
When compared to the most competitive work DeductReasoner, our model obtains a relative improvement of 3.84\%p on total benchmarks.

\input{figures/figure-addtraining.tex}

\paragraph{Performance on One-to-Many Test.} We note that ATHENA achieves large performance gains compared to the second-best method, from 42.5\% to 52.4\% and from 26.5\% to 35.0\% on SVAMP (1:N) and UnbiasedMWP (1:N), respectively.
As illustrated in Section~\ref{sec:one_to_many}, we evaluate our model on SVAMP (1:N) by training with one example per problem set to test how well ATHENA reasons on the questions that use the same textual descriptions but ask for different target answers.
We observe from Figure~\ref{fig:addtraining} that the results show ATHENA is strong at applying mathematical reasoning that is formed by unlearned patterns once the model has learned the context. Our approach is distinguished from other baselines including RoBERTa-GTS and DeductReasoner which show the opposite phenomena.
Other baselines are relatively stronger on original benchmarks than on the benchmark variants including those with the one-to-many Test.
Hence we reach the conclusion that ATHENA has the superiority of acknowledging the subtlety of contextual information governed by the required mathematical operations.

\begin{figure*}[t]
    \centering
    \includegraphics[width=\textwidth]{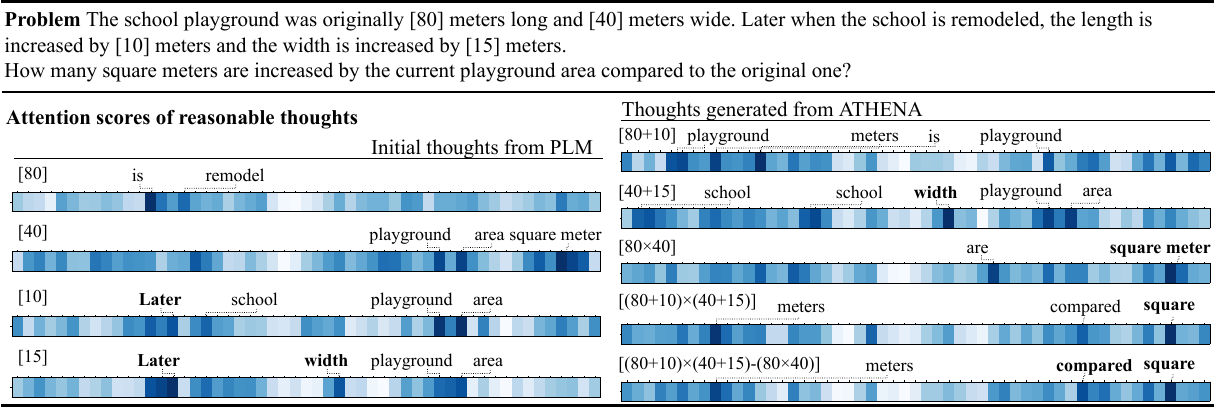}
    \vspace{-1.5em}
    \caption{Visualization of reasonable thoughts from ATHENA with calculating attention score of the tokens in the problem sequence on RoBERTa-large.}
    \label{fig:attention}
\end{figure*}

\input{figures/figure-organswer.tex}

\paragraph{Dependence on Training Set.}
We observe that ATHENA performs well on datasets that apply the one-to-many test because our model has a sense of subtlety in terms of distinct question concepts, not because our model is reluctant to follow learned expressions. Figure~\ref{fig:organswer} illustrates where the wrong prediction for the question variant experiments comes from.
If a model outputs equations that are labeled for questions with shared contexts when being trained, this indicates that the model relies on training data points, especially on context contents regardless of different question expressions.
The result shows that our model also has the least accuracy for a golden training example.
It is notable that ATHENA has the lowest score for following the trained expressions while DeductiveReasoner predicts the highest scores among other baselines that use RobBERTa, even higher than those of R-GTS or R-Graph-to-Tree on UnbiasedSVAMP(1:N).
This shows that while DeudctiveReasoner can learn to solve mathematical problems, it also easily falls into learning shortcuts.

\paragraph{Different Sizes of PLMs. }
We estimate the baselines both on RoBERTa-base and RoBERTa-large models to examine the influence of the model sizes. As expected, Table~\ref{tab:main} shows that the bigger the model size is for the embedding, the better the model performance reaches. When we estimate the accuracy gaps by increasing the model size, ATHENA achieves relatively better performance gains (7.26\%p) on average for the entire benchmarks than DeductReasoner does (4.6\%p). We can observe that on dataset variants, ATHENA obtains relatively more benefits from bigger model sizes (14.15\%p) than DeductReasoner does (8.15\%p), while both are still taking great advantage of the rich model parameters to understand the question better and to solve those confusing questions.
It also shows that DeductReasoner fails to improve performance on question variants from the original datasets leveraging the additional training sets in large-scale PLM.
In short, our model leverages large-scale PLM much more efficiently than the competitive model.

\input{figures/table-ablation.tex}

\paragraph{Visualization of Thoughts.}
We interpret the thoughts using attention scores between reasonable thoughts and the problem sequence.\footnote{We use $\answer$ layer to calculate the attention score, giving the problem sequence embedding as an input, instead of the goal vector.}
As illustrated in Figure~\ref{fig:attention}, we observe how the thought relates to the words.  
Most of the initial thoughts are related to the ``playground'', while the thoughts carrying the meaning of increased size show a strong correlation to the word ``Later''.
The thoughts carrying width sizes [15] and [40+15] show high attention scores on ``width'', while the other thoughts do not have high attention scores on them.
Thoughts that calculate the area produce high attention scores on words ``square meter'' or ``area''.
The final thought marks a high score on ``compared'', which asks for the difference between the increased and original areas.

\paragraph{Ablation on Premise Vectors.}
A premise vector is a criterion for determining thoughts in each inference step to obtain the valid pathways to reach the goal.
We conduct an ablation study to evaluate how ATHENA composes the premise vectors to ultimately generate optimal final thoughts.  

For evaluating the impact of the premise vectors in generating reasonable thoughts, we adopt two different settings:
(1) We do not \textit{update} premise vectors but use the initial premise vector (i.e., [CLS] token) in all expansion depths: $\PV_d=\PV_0$.
We aim to see how the existence of thoughts that update the premise vector impacts models to help find solid reasonable thoughts.
(2) We do not use the premise vector and directly classify the thoughts for the next iteration:  $\infer(\theta)=\sigma(\theta W_r+b_r)$.

Table~\ref{tab:reasoning} shows the results of the different premise construction strategies for reaching the appropriate conclusion.
Despite slight fluctuations across different methods, ATHENA without premise vectors decreases the overall performances by up to 3.7\%p compared to our proposed method.
When the model does not update the premise vectors in the thought expansion iteration while still adopting the initial one, the performance decreases relatively by 1.1\%p.
It is notable that Math23k, a dataset of the deepest average depth, shows the performance degradation even worse, 2.4\%p and 19.4\%p respectively.
From these observations, we conclude that the premise vector plays an important role in properly deriving the final thoughts.
Especially, considering that the model applies the update on every expansion depth, the large performance gap for Math23k strongly supports our premise update method for its effectiveness.

\section{Conclusion}
We state that an ideal MWP model needs to be practical in real-world settings that are critical to capture the diverse applications of the same mathematical operations.
For this reason, we conclude that ATHENA with thought expansion reaches significant improvements toward the ideal model due to its decent performance on unseen problems or restricted examples to learn.

\section*{Limitations}
The paper has the following limitations.
First, we only consider arithmetic problems, not algebraic, calculus, or other topics of mathematical problems.
Especially, for a fair comparison with other 
models, we only evaluate the performance using MWP datasets with single equations, while the model is able to handle multi-equation problems by simply adding ``='' operation on Merge.
Second, we do not compare ATHENA with large-scale language models (LLMs) since 
we focus on acquiring knowledge from limited mathematical samples.

\section*{Ethics Statement}
This work breaks down a process of reasoning from the human cognitive perspective and instantiates individual thoughts with symbolic representation so that it can clarify and handle the intermediate procedures of the model.
Although the perspective may have the potential to filter harmful or toxic thoughts from the broad sight of thoughts, this work does not consider or validate the effectiveness of such applications.
Therefore, we do not suggest using our work for this purpose without thorough experiments for its possibilities.

\section*{Acknowledgements}
This research was supported by the NRF grant (RS-2023-00208094) and the AI Graduate School Program (No.~2020-0-01361) funded by the Korean government~(MSIT).
Han is a corresponding author.
\bibliography{athena}
\bibliographystyle{acl_natbib}

\clearpage
\appendix

\section*{Appendices}
\section{Training Details}
In this section, we provide detailed information about our training settings.
\label{sec:appendix_training}

\paragraph{Loss.}
Given an answer equation $\Eq$, let $\1_{\infer}(\theta)$ denote the target of $\infer$ for a thought $\theta$ and $\1_{\answer}(\theta)$ denote the target of the final decision $\answer$ for a thought $\theta$:
\begin{align*}
\1_{\infer}(\theta) = \1(\Eq(\theta) \subseteq \Eq), \\
\1_{\answer}(\theta) = \1(\Eq(\theta) \equiv \Eq) ,
\end{align*}
where $\Eq(\theta) \subseteq \Eq$ denotes that $\Eq$ contains the sub-expression $\Eq(\theta)$ (e.g., ``$(a+b)$'' $\subseteq$ ``$(a+b)\times c$'').

Let $BCE$ denote the binary cross entropy function, the training objective is to minimize loss $\mathcal{L}$:
\begin{align*}
\mathcal{L}& = \frac{1}{|\bigcup_d \Theta_d|+|\Theta^*_d|}
\Big( \\
& \quad \sum\limits_{\theta\in \bigcup_d \Theta_d}BCE(\infer(\theta), \1_{\infer}(\theta)) \\
& \quad + \sum\limits_{\theta\in \Theta^*_d}BCE(\answer(\theta), \1_{\answer}(\theta))
\Big)
 .
\end{align*}

\paragraph{Optimizer.}
We use AdamW optimizer \citep{loshchilov-etal-2017-decoupled} with weight decay $\omega=10^{-5}$.
Learning rate $lr_e$ for each epoch $e$ is decayed every $S_{lr}$ epoch with factor $\gamma$
starting from $lr$:
\begin{equation*}
lr_e = lr \cdot \gamma^{ \left[e/S_{lr} \right] } .
\end{equation*}

\paragraph{Regularization.}
We adopt dropout with probability $p$ to every layer and
stochastic weight averaging \citep{izmailov-etal-2018-averaging} for last $epoch_{swa}$ epochs.

\paragraph{Hyperparameters.}
We present our experiments for hyperparameters in Table~\ref{tab:hyper},
with the bold text denoting the best performance.
We train our model for $100$ epochs.
In the result, we observe that RoBERTa-base and RoBERTa-large share the best hyperparameter settings except for learning rate $lr$.

\section{Dataset Statistics}
In this section, we show the statistics of datasets and their requirements.

\paragraph{One-to-many Split.}
In Section~\ref{sec:one_to_many}, we explain building one-to-many dataset splits.
We provide how many groups and examples are made from the contexts in Table~\ref{tab:otm_statistics}.

\paragraph{Number of Thoughts.}
We present the required number of thoughts for each dataset in Table~\ref{tab:thoughts}.
While Math23k requires a large number of candidate thoughts in total depth, we show a thought expansion in each depth does not require huge memory space. Therefore, efficient implementation strategies such as removing unselected candidate thoughts from memory space are enough to manage computational resources.

\section{Additional Experiments}
This section presents additional studies to further clarify the robustness and fairness of our experiments for some minor strategies by showing their performance independence.

\paragraph{Punctuation Mark.}
In Section~\ref{sec:athena}, we initialize goal vector $\GV$ with the punctuation mark of the question sequence or the last punctuation mark (i.e., the question mark in most cases).
The motivation of this strategy is from \citet{clark-etal-2019-bert} showing the punctuation mark gets high attention from other tokens in the last layers.
Intuitively, high attention can generalize the question sequence,
so we conduct experiments to evaluate the generalization ability of the punctuation mark compared to using all question sequences as a goal vector $\GV$.
We conduct experiments for all datasets except Math23k~\citep{wang-etal-2017-deep} since it does not provide the question subsequence.

As shown in Table~\ref{tab:question}, using the punctuation mark effectively generalizes the question to represent a goal in most cases.
It shows even better performances than using the question sequence.
From an intuitive interpretation, the question sequence holds some tokens that are not informative for reasoning targets, so a punctuation mark representation helps the model to focus on a reasoning goal.

\paragraph{Stop Criteria.}
In Section~\ref{sec:final} and Algorithm~\ref{alg:athena}, we present the two stop criteria: (1) the depth reaches the maximum expansion depth $D$, or (2) one of the final scores exceeds a threshold $t_f$.
In addition to the main experiments setting the $D$ and $t_f$ with an arbitrary value,
we conduct the experiments of the higher maximum expansion depth and $t_f=0.5$ to show the performance differences from the values.
As shown in Table~\ref{tab:stop}, the scores are fairly equal with a trivial gap.
This demonstrates that the performances of our model do not rely on stop criteria parameters but are solid achievements.

\begin{table*}
    \centering
    \setlength{\tabcolsep}{3pt}
    \small
    \begin{tabular}{lcccccc}
        \toprule
        & Batch Size & $lr$ & $S_{lr}$ & $\gamma$ & $p$ & $epoch_{swa}$ \\
        \midrule
        RoBERTa-base & [\textbf{4}, 8]  & [5e-6, 7e-6, 1e-5, \textbf{1.3e-5}, 1.5e-5, 2e-5] & [\textbf{10}, 15, 20] & [\textbf{0.5}, 0.7] & [0.1, \textbf{0.5}] & [\textbf{30}, 50, 70] \\
        RoBERTa-large & [\textbf{4}, 8]  & [5e-6, \textbf{7e-6}, 1e-5, 1.3e-5, 1.5e-5, 2e-5] & [\textbf{10}, 15, 20] & [\textbf{0.5}, 0.7] & [0.1, \textbf{0.5}] & [\textbf{30}, 50, 70] \\
         \bottomrule
    \end{tabular}
    \caption{Hyperparameter search spaces of ATHENA}
    \label{tab:hyper}
\end{table*}

\begin{table*}
    \centering
    \small
    \begin{tabular}{lcc}
        \toprule
         & \textbf{SVAMP (1:N)} & \textbf{UnbiasedMWP (1:N)} \\
        \midrule
        \# examples in original split  & 3138 / 0 / 1000 & 2507 / 200 / 685 \\
        \# groups of single examples & 438 & 45 \\
        \# groups of multiple examples & 205 & 154\\
        \# examples in one-to-many split & 3343 (+205) / 438 (+438) / 357 (-562) & 2661 (+154) / 245 (+45) / 486 (-199) \\
        \bottomrule
    \end{tabular}
    \caption{Statistics of one-to-many test splits}
    \label{tab:otm_statistics}
\end{table*}

\begin{table*}
    \centering
    \setlength{\tabcolsep}{5pt}
    \small
    \begin{tabular}{l|ccc|ccc|ccc|ccc}
        \toprule
        \textbf{Dataset}
        & \multicolumn{3}{|c|}{\# candidates in total depth} & \multicolumn{3}{c|}{\# in a reasoning path}
        & \multicolumn{3}{c|}{\# candidates in last depth} & \multicolumn{3}{c}{depth of reasoning path} \\
        \addlinespace[0.4ex]
        \midrule
        \addlinespace[0.4ex]
        & min & average & max & min & average & max & min & average & max & min & average & max\\[0.65ex]
        MAWPS & 17 & 45.40\textpm0.46 & 192 & 2 & 4.52\textpm0.03 & 12  & 4 & 9.49\textpm0.08 & 48 & 2 & 3.87\textpm0.03 & 11\\[.4ex]
        ASDiv-A & 16 & 26.86\textpm0.42 & 71 & 3 & 4.10\textpm0.03 & 7 & 6 & 9.65\textpm0.09 & 22 & 1 & 3.46\textpm0.02 & 5\\[.4ex]
        SVAMP & 2 & 28.09\textpm0.44 & 70 & 1 & 4.23\textpm0.03 & 7 & 2 & 10.54\textpm0.10 & 22 & 1 & 3.47\textpm0.03 & 5\\[.4ex]
        Math23k & 4 & 65.1\textpm0.31 & 939 & 1 & 6.33\textpm0.02 & 29 & 2 & 14.85\textpm0.06 & 108 & 1 & 5.18\textpm0.01 & 41\\[.4ex]
        U.MWP & 5 & 47.0\textpm0.47 & 214 & 1 & 5.18\textpm0.03 & 13 & 2 & 11.67\textpm0.11 & 48 & 1 & 4.44\textpm0.02 & 11\\
        \bottomrule
    \end{tabular}
    \caption{Statistics of thoughts that are required for each dataset}
    \label{tab:thoughts}
\end{table*}

\begin{table*}
    \centering
    \setlength{\tabcolsep}{2pt}
    \small
    \begin{tabular}{lcccccc|c}
        \toprule
         & \textbf{MAWPS} & \textbf{ASDiv-A} & \textbf{SVAMP} &  \textbf{UnbiasedMWP} & \textbf{SVAMP (1:N)} & \textbf{UnbiasedMWP (1:N)} & \textbf{Average}\\[-.5ex]
         \multicolumn{1}{c}{\tiny{Avg depth}} & \tiny{3.87} & \tiny{3.46} & \tiny{3.47} & \tiny{4.44} &  \tiny{3.47} & \tiny{4.44} & \tiny{4.05} \\
         \midrule
         {\tiny RoBERTa-base} \\
         punctuation mark  & \textbf{92.2} & \textbf{86.4} & \textbf{45.6} 
         & 36.2  & \textbf{52.5} & \textbf{35.4} & \textbf{58.1} \\
        question sequence & 92.0 & 86.3 & 44.9  & \textbf{36.3} & 51.0 & 33.4 & 57.3 \\
        \midrule
        {\tiny RoBERTa-large} \\
        punctuation mark & \textbf{93.0} & 91.0 & \textbf{54.8}
          & \textbf{42.0} & \textbf{67.8} & \textbf{48.4} & \textbf{66.2} \\
          question sequence & 92.9 & \textbf{91.2} & 54.4  & 41.0 & 66.9 & 46.8 & 65.5 \\
        \bottomrule
    \end{tabular}
    \caption{Comparing goal vector using the whole question sequence from the punctuation mark}
    \label{tab:question}
\end{table*}

\begin{table*}
    \centering
    \setlength{\tabcolsep}{2pt}
    \small
    \begin{tabular}{lccccc|c}
        \toprule
         \textbf{Stop Criteria} & \textbf{MAWPS} & \textbf{ASDiv-A} & \textbf{SVAMP} & \textbf{Math23k} & \textbf{UnbiasedMWP} & \textbf{Average}\\[-.5ex]
          & \tiny{D=7} & \tiny{D=5} & \tiny{D=5} & \tiny{D=19} &  \tiny{D=9} &  \\
         \midrule
         $t_f=0.95$, max. expansion depth=$D$ & \textbf{92.2{\tiny\textpm0.10}} & 86.4{\tiny\textpm0.11} & \textbf{45.6{\tiny\textpm0.50}} & 84.4{\tiny\textpm0.24} & 36.2{\tiny\textpm0.67} & \textbf{69.0} \\
         \midrule
         $t_f=0.5$, max. expansion depth=$D$ & 92.0{\tiny\textpm0.15} & 86.3{\tiny\textpm0.24} & 45.3{\tiny\textpm0.37} & 84.7{\tiny\textpm0.20} & 36.0{\tiny\textpm0.39} & 68.9 \\
         $t_f=0.95$, max. expansion depth=$D+2$ & 91.9{\tiny\textpm0.07} & \textbf{86.5{\tiny\textpm0.28}} & 45.2{\tiny\textpm0.41} & 84.6{\tiny\textpm0.18} & 35.8{\tiny\textpm0.75} & 68.8 \\
         $t_f=0.95$, max. expansion depth=$D+4$ & 92.0{\tiny\textpm0.09} & \textbf{86.5{\tiny\textpm0.28}} & 45.2{\tiny\textpm0.41} & \textbf{84.8{\tiny\textpm0.27}} & \textbf{36.4{\tiny\textpm0.44}} & \textbf{69.0} \\
        \midrule
    \end{tabular}
    \caption{Performances among the different parameters of stop criteria}
    \label{tab:stop}
\end{table*}

\end{document}

%% file: figures/table-dr-athena.tex
 \begin{table*}
    \centering
    \small
    \setlength{\tabcolsep}{0pt}
    \begin{tabular}{p{.55\textwidth}p{.45\textwidth}}
        \toprule
        \multicolumn{2}{p{\textwidth}}{
        \textbf{Context} The school playground was originally [80] meters long and [40] meters wide. Later when the school is remodeled, the length is increased by [10] meters and the width is increased by [15] meters.
        } \\
        \midrule
        \multicolumn{2}{c}{
        \scriptsize Train on an example of a question-solution pair under the context above.
        }\\[.5ex]
        \textbf{Question} How many square meters is the original playground area? 
        & \hspace{1em} \textbf{Solution} $(80 \times 40)$ \\
        
        \midrule
        \multicolumn{2}{c}{
        \scriptsize Test on variant questions that share the context above.
        }\\[.5ex]
        \multicolumn{2}{p{\textwidth}}{
        \textbf{Q0} How many times the length of the original playground was the width?
        } \\[.7ex]
        \textbf{DeductReasoner} & \hspace{1em} \textbf{ATHENA} \\
        {\scriptsize UnbiasedMWP \textcolor{red}{$(80 + 10) \times (40 + 15) - (80 \times 40)$} } (X) 
        & \hspace{1em} {\scriptsize UnbiasedMWP \textcolor{red}{$ (80 + 10) \times (40 - 15) $} } (X) \\
        {\scriptsize UnbiasedMWP (1:N) \textcolor{blue}{$80 \div 40$} } (O)
        &\hspace{1em}  {\scriptsize UnbiasedMWP (1:N) \textcolor{blue}{$80 \div 40$}} (O) \\

        \midrule[.1pt]

        \multicolumn{2}{p{\textwidth}}{
        \textbf{Q1} How many square meters is the current playground area?
        } \\[.7ex]
		\textbf{DeductReasoner} & \hspace{1em} \textbf{ATHENA} \\
        {\scriptsize UnbiasedMWP \textcolor{red}{$(80 + 10) \times (40 + 15) - (80 \times 40)$} } (X)
        & \hspace{1em} {\scriptsize UnbiasedMWP \textcolor{blue}{$ (80 + 10) \times (40 + 15) $} } (O)\\
        {\scriptsize UnbiasedMWP (1:N) \textcolor{red}{$80 \times 40$}} (X)
        & \hspace{1em} {\scriptsize UnbiasedMWP (1:N) \textcolor{blue}{$ (80 + 10) \times (40 + 15) $} } (O)\\

        \midrule[.1pt]
        \multicolumn{2}{p{\textwidth}}{
        \textbf{Q2} How many square meters are increased by the current playground area compared to the original one?
		} \\[.7ex]
		\textbf{DeductReasoner} & \hspace{1em} \textbf{ATHENA} \\
        {\scriptsize UnbiasedMWP \textcolor{blue}{$(80 + 10) \times (40 + 15) - (80 \times 40)$}} (O)
        & \hspace{1em} {\scriptsize UnbiasedMWP \textcolor{blue}{$ (80 + 10) \times (40 + 15) - (80 \times 40) $} } (O)\\
        {\scriptsize UnbiasedMWP (1:N) \textcolor{red}{$80 \times 40$} } (X)
        & \hspace{1em} {\scriptsize UnbiasedMWP (1:N) \textcolor{blue}{$ (80 + 10) \times (40 + 15) - (80 \times 40) $} } (O)\\

        \midrule
        \multicolumn{2}{c}{
        \scriptsize An example with a lexically similar context to that of above from the UnbiasedMWP}\\[.5ex]
        \multicolumn{2}{p{\textwidth}}{
        \textbf{Context} The school basketball court was [20] meters long and [12] meters wide. After the renovation, the length is increased by [8] meters, and the width increases by [3] meters.
        }\\
        \textbf{Question} How many square meters are increased? & \hspace{1em} \textbf{Solution} $(20+8)\times(12+3)-(20\times12)$ \\

         \bottomrule
    \end{tabular}
    \caption{Predictions of DeductReasoner~\citep{jie-etal-2022-learning} and ATHENA on a sample that has variant questions while sharing the common context for the problems. The observation above is when models use RoBERTa-large on UnbiasedMWP~\citep{yang-etal-2022-unbiased}.}
    \label{tab:dr_athena}
    \vspace{-.1em}
\end{table*}

%% file: figures/table-main.tex
\begin{table*}[t]
    \centering
    \setlength{\tabcolsep}{4pt}
    \small
    \begin{tabular}{lccc|cc|cc}
        \toprule
         & \textbf{MAWPS} & \textbf{ASDiv-A} &  \textbf{Math23k} &\textbf{SVAMP} & \textbf{UnbiasedMWP} & \textbf{SVAMP (1:N)} & \textbf{UnbiasedMWP (1:N)}\\[-.5ex]
        \multicolumn{1}{r}{\tiny{Language}} & \tiny{English} & \tiny{English} &  \tiny{Chinese} & \tiny{English} & \tiny{Chinese} &  \tiny{English} & \tiny{Chinese}  \\
        \bottomrule[0.01em]
        {\tiny Random embedding} \\
        Transformer     & 85.6 & 59.3 & 61.5 & 20.7 & 20.5{\tiny\textpm0.73} & 9.7{\tiny\textpm0.19} (14.9)  & 16.9{\tiny\textpm0.31} (51.5)\\
        GTS             & 82.6 & 71.4 & 75.6 & 30.8 & 26.2{\tiny\textpm0.20} & 12.2{\tiny\textpm0.37} (43.8) & 22.8{\tiny\textpm0.22} (65.0)\\
        Graph-to-Tree   & 83.7 & 77.4 & 77.4 & 36.5 & 27.2{\tiny\textpm0.37} & 25.3{\tiny\textpm0.12} (52.5) & 24.3{\tiny\textpm0.25} (66.4)\\
        \bottomrule[0.01em]
        {\tiny RoBERTa-base} \\
        R-Transformer     & 88.4 & 72.1 & 76.9 & 30.3 &   18.3{\tiny\textpm0.15}    & 13.5{\tiny\textpm0.33} (33.4) & 14.9{\tiny\textpm0.20} (53.1) \\
        R-GTS             & 88.5 & 81.2 & -    & 41.0 &  - & 40.9{\tiny\textpm0.50} (64.4) & - \\
        R-Graph-to-Tree   & 88.7 & 82.2 & -    & 43.8 &  - & 31.8{\tiny\textpm0.36} (66.7) & - \\
        DeductReasoner  & 92.0{\tiny\textpm0.20} & 83.1{\tiny\textpm0.24}
        & \textbf{85.1}{\tiny\textpm0.24}  & 45.0{\tiny\textpm0.10} & 31.6{\tiny\textpm0.51} & 42.5{\tiny\textpm0.41} (69.1) & 26.5{\tiny\textpm0.55} (79.5) \\[.65ex]
        \textbf{ATHENA}(Ours)  
        & \textbf{92.2}{\tiny\textpm0.10} & \textbf{86.4}{\tiny\textpm0.11} &  84.4{\tiny\textpm0.24} & \textbf{45.6}{\tiny\textpm0.50} & \textbf{36.2}{\tiny\textpm0.67} & \textbf{52.5}{\tiny\textpm0.50} (70.1) & \textbf{35.4}{\tiny\textpm0.45} (80.5) \\
        \bottomrule[0.01em]
        {\tiny RoBERTa-large} \\
        DeductReasoner  & 92.6{\tiny\textpm0.16} & 89.1{\tiny\textpm0.46}
        & 85.8{\tiny\textpm0.42}  & 50.3{\tiny\textpm0.30} & 34.9{\tiny\textpm0.11}  & 51.6{\tiny\textpm0.38} (75.4)   & 33.7{\tiny\textpm0.60} (83.2) \\[.65ex]
        \textbf{ATHENA}(Ours)
        & \textbf{93.0}{\tiny\textpm0.20} & \textbf{91.0}{\tiny\textpm0.13} & \textbf{86.5}{\tiny\textpm0.25}  & \textbf{54.8}{\tiny\textpm0.63} & \textbf{42.0}{\tiny\textpm0.57} & \textbf{67.8}{\tiny\textpm0.58} (79.8) & \textbf{48.4}{\tiny\textpm0.38} (84.8)\\
        \bottomrule
    \end{tabular}
    \caption{Comparison of MWP methods. We use MAWPS, ASDiv-A, and Math23k for standard evaluation, SVAMP and UnbiasedMWP to evaluate the ability to solve entirely unseen, various expressions, and  SVAMP and UnbiasedMWP with the one-to-many test to estimate the adaptability of confusing linguistic subtlety.}
    \label{tab:main}
\end{table*}

%% file: figures/figure-addtraining.tex
\pgfplotstableread{
model            not1  add1 neg1  nmeta1   ameta1   not2  add2 ameta2
Transformer       9.7  0    0.8   10.5    {-0.8}    15.6 1.3 {+1.3}
GTS              12.2  0    3.0   15.2    {-3.0}    21.6 0.8 {+0.8}
Graph-to-Tree    18.3  7.0   0    18.3    {+7.0}    22.4 1.9 {+1.9}
R-Transformer    13.5  0    2.0   15.5    {-2.0}    13.5 1.4 {+1.4}
DeductReasoner   25.0  17.5  0    25.0    {+17.5}   24.5 2.0 {+2.0}
ATHENA           27.7  24.8  0    27.7    {+24.8}   28.5 6.9 {+6.9}
}\addtraindata

\begin{figure}[t]
    \centering

    \begin{tikzpicture}[scale=0.9]
        \small
        \begin{axis}[
                width=\columnwidth,
                height=.8\columnwidth,
                xbar stacked,
                bar width=0.3cm,
                xmin=0,
                symbolic y coords={ATHENA,DeductReasoner,R-Transformer,Graph-to-Tree,GTS,Transformer},
                ytick distance=1,
                xticklabel={\pgfmathparse{\tick}\pgfmathprintnumber{\pgfmathresult}\%},
                point meta=explicit symbolic,
                nodes near coords,
                nodes near coords style={/pgf/number format/precision=1, color=black,font=\tiny},
                legend image code/.code={ \draw[#1,draw=black] (0cm,-0.1cm) rectangle (0.2cm,0.1cm); },
                legend style={font=\tiny},
                axis x line*=bottom,
                axis y line*=left,
                ytick style={draw=none},
                xtick style={draw=none},
            ]
            \addplot +[fill=blue!10,draw=none,bar shift=.17cm,nodes near coords align={center}] table [meta=nmeta1,y=model,x=not1] {\addtraindata};
            \addplot +[fill=blue!30,draw=none,bar shift=.17cm,nodes near coords align={center}] table [meta=ameta1,y=model,x=add1] {\addtraindata};
            \addplot +[fill=red!30,draw=none,bar shift=.17cm,nodes near coords align={center}] table [meta=ameta1,y=model,x=neg1] {\addtraindata};

\makeatletter\pgfplots@stacked@isfirstplottrue\makeatother
            \addplot +[fill=green!10,draw=none,bar width=0.3cm,bar shift=-.17cm,
            ] table [meta=not2,y=model,x=not2] {\addtraindata};
            \addplot +[fill=green!30,draw=none,bar shift=-.17cm] table [meta=ameta2,y=model,x=add2] {\addtraindata};
            \legend{SVAMP (1:N) w/o adding,SVAMP (1:N),UnbiasedMWP (1:N) w/o adding,UnbiasedMWP (1:N)}
        \end{axis}
    \end{tikzpicture}
    \caption{Accuracy changes when adding one example per context into the training set by applying the one-to-many test.}
    \label{fig:addtraining}
\end{figure}

%% file: figures/figure-organswer.tex
\pgfplotstableread{
model            oa1      wrong1   oa2      wrong2  all1 all2 acc1   acc2
GTS              199.6    302.8    138.6    338.4   339  439  {}   {}
Graph-to-Tree    136.2    258.8    131.2    327.6   339  439  {}   {}
R-Transformer    164.0    294.2    162.2    374.6   339  439  {}   {}
R-Graph-to-Tree  111.08   241.12   nan      0    339  439  {}  n/a 
R-GTS            116.64   205.64   nan      0    339  439  {}  n/a
DeductReasoner   131.6    201.6    172.36   321.24  339  439  {}   {}
ATHENA           85.7     164.68   96.44    285.44  339  439  {}   {}
}\oadata
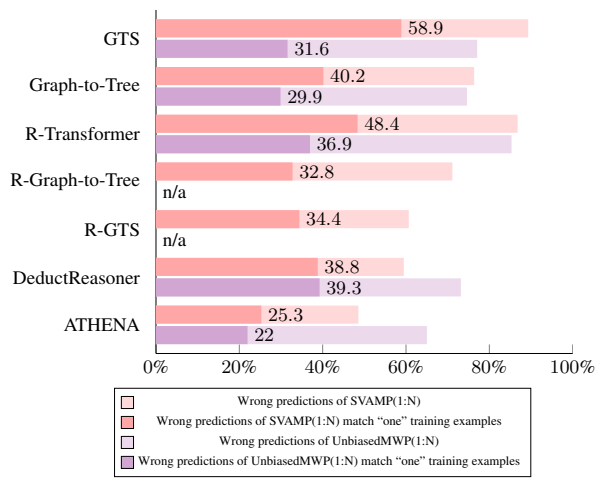
\begin{figure}[t]
    \centering

    \begin{tikzpicture}[scale=0.8]
        \small
        \begin{axis}[
                xbar,
                bar width=0.3cm,
                xmin=0,
                xmax=100,
                symbolic y coords={ATHENA,DeductReasoner,R-GTS,R-Graph-to-Tree,R-Transformer,Graph-to-Tree,GTS},
                ytick distance=1,
                xticklabel={\pgfmathparse{\tick}\pgfmathprintnumber{\pgfmathresult}\%},
                point meta=explicit symbolic,
                nodes near coords,
                nodes near coords style={/pgf/number format/precision=1, /pgf/number format/fixed, color=black},
                nodes near coords align={right},
                legend image code/.code={ \draw[#1,draw=black] (0cm,-0.1cm) rectangle (0.2cm,0.1cm); },
                legend style={at={(0.4,-.1)},anchor=north,font=\tiny},
                axis x line*=bottom,
                axis y line*=left,
                ytick style={draw=none},
            ]
            \addplot +[bar shift=.17cm,draw=none,fill=red!15] table [meta=acc1,y=model,x expr=\thisrow{wrong1}/3.39] {\oadata};
            \addplot +[bar shift=.17cm,draw=none,fill=red!35,point meta=x] table [y=model,x expr=\thisrow{oa1}/3.39] {\oadata};
            \addplot +[bar shift=-.17cm,draw=none,fill=violet!15] table [meta=acc2,y=model,x expr=\thisrow{wrong2}/4.39] {\oadata};
            \addplot +[bar shift=-.17cm,draw=none,fill=violet!35,point meta=x] table [y=model,x expr=\thisrow{oa2}/4.39] {\oadata};
            \legend{Wrong predictions of SVAMP(1:N)
,Wrong predictions of SVAMP(1:N) match ``one'' training examples,Wrong predictions of UnbiasedMWP(1:N),Wrong predictions of UnbiasedMWP(1:N) match ``one'' training examples}
        \end{axis}
    \end{tikzpicture}
    \caption{Percentages of the wrong predictions that match with the answers of ``one'' training examples in one-to-many split. The less the percentage scores, the less the method unnecessarily leans on the training bias.
    }
    \label{fig:organswer}
\end{figure}

%% file: figures/table-ablation.tex
\begin{table*}[t]
    \centering
    \setlength{\tabcolsep}{2pt}
    \small
    \begin{tabular}{lccccccc|c}
        \toprule
         & \textbf{MAWPS} & \textbf{ASDiv-A} & \textbf{SVAMP} & \textbf{Math23k} & \textbf{UnbiasedMWP} & \textbf{SVAMP (1:N)} & \textbf{UnbiasedMWP (1:N)} & \textbf{Average}\\[-.5ex]
         \multicolumn{1}{c}{\tiny{Avg depth}} & \tiny{3.87} & \tiny{3.46} & \tiny{3.47} & \tiny{5.18} & \tiny{4.44} &  \tiny{3.47} & \tiny{4.44} & \tiny{4.05} \\
         \midrule
         \textbf{ATHENA}  & \textbf{92.2} & \textbf{86.4} & \textbf{45.6} 
        & \textbf{85.1} & 36.2  & \textbf{52.5} & \textbf{35.4} & \textbf{62.0} \\
        {\scriptsize $-$ update} & 92.1 & 84.8 & 44.9 & 82.7 & 34.9 & 52.4 & 34.7 & 60.9 \\
        {\scriptsize $-$ premise} &90.6&85.0&44.7& 65.7 & \textbf{36.3} & 51.5 & 34.6 & 58.3 \\
        \bottomrule
    \end{tabular}
    \caption{Ablation studies on premise vector construction. (1) ``$-$ update'' is the premise vector without updating strategies and (2) ``$-$ premise'' is the direct classification method without premise vectors.}
    \label{tab:reasoning}
\end{table*}

%% file: athena.bbl
\begin{thebibliography}{42}
\expandafter\ifx\csname natexlab\endcsname\relax\def\natexlab#1{#1}\fi

\bibitem[{Bakman(2007)}]{bakman-etal-2007-robust}
Yefim Bakman. 2007.
\newblock \href {https://doi.org/10.48550/arXiv.math/0701393} {Robust understanding of word problems with extraneous information}.
\newblock \emph{arXiv preprint math/0701393}.

\bibitem[{Byrnes and Wasik(1991)}]{byrnes1991role}
James~P Byrnes and Barbara~A Wasik. 1991.
\newblock Role of conceptual knowledge in mathematical procedural learning.
\newblock \emph{Developmental psychology}, 27(5):777.

\bibitem[{Canobi(2009)}]{canobi2009concept}
Katherine~H Canobi. 2009.
\newblock Concept--procedure interactions in children’s addition and subtraction.
\newblock \emph{Journal of experimental child psychology}, 102(2):131--149.

\bibitem[{Chiang and Chen(2019)}]{chiang-chen-2019-semantically}
Ting-Rui Chiang and Yun-Nung Chen. 2019.
\newblock \href {https://doi.org/10.18653/v1/N19-1272} {Semantically-aligned equation generation for solving and reasoning math word problems}.
\newblock In \emph{Proceedings of the 2019 Conference of the North {A}merican Chapter of the Association for Computational Linguistics: Human Language Technologies, Volume 1 (Long and Short Papers)}, pages 2656--2668.

\bibitem[{Clark et~al.(2019)Clark, Khandelwal, Levy, and Manning}]{clark-etal-2019-bert}
Kevin Clark, Urvashi Khandelwal, Omer Levy, and Christopher~D. Manning. 2019.
\newblock \href {https://doi.org/10.18653/v1/W19-4828} {What does {BERT} look at? an analysis of {BERT}{'}s attention}.
\newblock In \emph{Proceedings of the 2019 ACL Workshop BlackboxNLP: Analyzing and Interpreting Neural Networks for NLP}, pages 276--286.

\bibitem[{Cui et~al.(2019)Cui, Che, Liu, Qin, Yang, Wang, and Hu}]{chinese-roberta}
Yiming Cui, Wanxiang Che, Ting Liu, Bing Qin, Ziqing Yang, Shijin Wang, and Guoping Hu. 2019.
\newblock Pre-training with whole word masking for chinese bert.
\newblock \emph{arXiv preprint arXiv:1906.08101}.

\bibitem[{Hosseini et~al.(2014)Hosseini, Hajishirzi, Etzioni, and Kushman}]{hosseini-etal-2014-learning}
Mohammad~Javad Hosseini, Hannaneh Hajishirzi, Oren Etzioni, and Nate Kushman. 2014.
\newblock \href {https://doi.org/10.3115/v1/D14-1058} {Learning to solve arithmetic word problems with verb categorization}.
\newblock In \emph{Proceedings of the 2014 Conference on Empirical Methods in Natural Language Processing ({EMNLP})}, pages 523--533.

\bibitem[{Huang et~al.(2021)Huang, Lin, Wang, Liu, Chen, Ma, Su, and Tong}]{huang-etal-2021-disenqnet}
Zhenya Huang, Xin Lin, Hao Wang, Qi~Liu, Enhong Chen, Jianhui Ma, Yu~Su, and Wei Tong. 2021.
\newblock \href {https://doi.org/10.1145/3447548.3467347} {Disenqnet: Disentangled representation learning for educational questions}.
\newblock In \emph{Proceedings of the 27th ACM SIGKDD Conference on Knowledge Discovery \& Data Mining}, page 696–704.

\bibitem[{Huang et~al.(2020)Huang, Liu, Gao, Wu, Yin, Wang, and Chen}]{huang-etal-2020-neural}
Zhenya Huang, Qi~Liu, Weibo Gao, Jinze Wu, Yu~Yin, Hao Wang, and Enhong Chen. 2020.
\newblock \href {https://doi.org/10.1145/3397271.3401227} {Neural mathematical solver with enhanced formula structure}.
\newblock In \emph{Proceedings of the 43rd International ACM SIGIR Conference on Research and Development in Information Retrieval}, page 1729–1732.

\bibitem[{Izmailov et~al.(2018)Izmailov, Podoprikhin, Garipov, Vetrov, and Wilson}]{izmailov-etal-2018-averaging}
Pavel Izmailov, Dmitrii Podoprikhin, Timur Garipov, Dmitry Vetrov, and Andrew~Gordon Wilson. 2018.
\newblock Averaging weights leads to wider optima and better generalization.
\newblock \emph{arXiv preprint arXiv:1803.05407}.

\bibitem[{Jie et~al.(2022)Jie, Li, and Lu}]{jie-etal-2022-learning}
Zhanming Jie, Jierui Li, and Wei Lu. 2022.
\newblock \href {https://doi.org/10.18653/v1/2022.acl-long.410} {Learning to reason deductively: Math word problem solving as complex relation extraction}.
\newblock In \emph{Proceedings of the 60th Annual Meeting of the Association for Computational Linguistics (Volume 1: Long Papers)}, pages 5944--5955.

\bibitem[{Johnson-Laird(2008)}]{johnson2008we}
Philip Johnson-Laird. 2008.
\newblock \emph{How we reason}.
\newblock Oxford University Press.

\bibitem[{Koncel-Kedziorski et~al.(2016)Koncel-Kedziorski, Roy, Amini, Kushman, and Hajishirzi}]{koncel-kedziorski-etal-2016-mawps}
Rik Koncel-Kedziorski, Subhro Roy, Aida Amini, Nate Kushman, and Hannaneh Hajishirzi. 2016.
\newblock \href {https://doi.org/10.18653/v1/N16-1136} {{MAWPS}: A math word problem repository}.
\newblock In \emph{Proceedings of the 2016 Conference of the North {A}merican Chapter of the Association for Computational Linguistics: Human Language Technologies}, pages 1152--1157.

\bibitem[{Lan et~al.(2022)Lan, Wang, Zhang, Lan, Dai, Wang, Zhang, and Lim}]{lan-etal-2022-mwptoolkit}
Yihuai Lan, Lei Wang, Qiyuan Zhang, Yunshi Lan, Bing~Tian Dai, Yan Wang, Dongxiang Zhang, and Ee-Peng Lim. 2022.
\newblock Mwptoolkit: An open-source framework for deep learning-based math word problem solvers.
\newblock \emph{Proceedings of the AAAI Conference on Artificial Intelligence}, 36(11):13188--13190.

\bibitem[{Li et~al.(2019)Li, Wang, Zhang, Wang, Dai, and Zhang}]{li-etal-2019-modeling}
Jierui Li, Lei Wang, Jipeng Zhang, Yan Wang, Bing~Tian Dai, and Dongxiang Zhang. 2019.
\newblock \href {https://doi.org/10.18653/v1/P19-1619} {Modeling intra-relation in math word problems with different functional multi-head attentions}.
\newblock In \emph{Proceedings of the 57th Annual Meeting of the Association for Computational Linguistics}, pages 6162--6167.

\bibitem[{Li et~al.(2020)Li, Wu, Feng, Xu, Xu, and Zhong}]{li-etal-2020-graph-tree}
Shucheng Li, Lingfei Wu, Shiwei Feng, Fangli Xu, Fengyuan Xu, and Sheng Zhong. 2020.
\newblock \href {https://doi.org/10.18653/v1/2020.findings-emnlp.255} {Graph-to-tree neural networks for learning structured input-output translation with applications to semantic parsing and math word problem}.
\newblock In \emph{Findings of the Association for Computational Linguistics: EMNLP 2020}, pages 2841--2852.

\bibitem[{Li et~al.(2022)Li, Zhang, Yan, Zhou, Li, Liu, and Cao}]{li-etal-2022-seeking}
Zhongli Li, Wenxuan Zhang, Chao Yan, Qingyu Zhou, Chao Li, Hongzhi Liu, and Yunbo Cao. 2022.
\newblock \href {https://doi.org/10.18653/v1/2022.findings-acl.195} {Seeking patterns, not just memorizing procedures: Contrastive learning for solving math word problems}.
\newblock In \emph{Findings of the Association for Computational Linguistics: ACL 2022}, pages 2486--2496.

\bibitem[{Liang et~al.(2022)Liang, Zhang, Wang, Qin, Lan, Shao, and Zhang}]{liang-etal-2022-mwp}
Zhenwen Liang, Jipeng Zhang, Lei Wang, Wei Qin, Yunshi Lan, Jie Shao, and Xiangliang Zhang. 2022.
\newblock \href {https://doi.org/10.18653/v1/2022.findings-naacl.74} {{MWP}-{BERT}: Numeracy-augmented pre-training for math word problem solving}.
\newblock In \emph{Findings of the Association for Computational Linguistics: NAACL 2022}, pages 997--1009.

\bibitem[{Lin et~al.(2021)Lin, Huang, Zhao, Chen, Liu, Wang, and Wang}]{lin-etal-2021-hms}
Xin Lin, Zhenya Huang, Hongke Zhao, Enhong Chen, Qi~Liu, Hao Wang, and Shijin Wang. 2021.
\newblock Hms: A hierarchical solver with dependency-enhanced understanding for math word problem.
\newblock \emph{Proceedings of the AAAI Conference on Artificial Intelligence}, 35(5):4232--4240.

\bibitem[{Ling et~al.(2017)Ling, Yogatama, Dyer, and Blunsom}]{ling-etal-2017-program}
Wang Ling, Dani Yogatama, Chris Dyer, and Phil Blunsom. 2017.
\newblock \href {https://doi.org/10.18653/v1/P17-1015} {Program induction by rationale generation: Learning to solve and explain algebraic word problems}.
\newblock In \emph{Proceedings of the 55th Annual Meeting of the Association for Computational Linguistics (Volume 1: Long Papers)}, pages 158--167.

\bibitem[{Liu et~al.(2019{\natexlab{a}})Liu, Guan, Li, and Kawahara}]{liu-etal-2019-tree}
Qianying Liu, Wenyv Guan, Sujian Li, and Daisuke Kawahara. 2019{\natexlab{a}}.
\newblock \href {https://doi.org/10.18653/v1/D19-1241} {Tree-structured decoding for solving math word problems}.
\newblock In \emph{Proceedings of the 2019 Conference on Empirical Methods in Natural Language Processing and the 9th International Joint Conference on Natural Language Processing (EMNLP-IJCNLP)}, pages 2370--2379.

\bibitem[{Liu et~al.(2019{\natexlab{b}})Liu, Ott, Goyal, Du, Joshi, Chen, Levy, Lewis, Zettlemoyer, and Stoyanov}]{liu-etal-2019-roberta}
Yinhan Liu, Myle Ott, Naman Goyal, Jingfei Du, Mandar Joshi, Danqi Chen, Omer Levy, Mike Lewis, Luke Zettlemoyer, and Veselin Stoyanov. 2019{\natexlab{b}}.
\newblock Roberta: A robustly optimized bert pretraining approach.
\newblock \emph{arXiv preprint arXiv:1907.11692}.

\bibitem[{Loshchilov and Hutter(2017)}]{loshchilov-etal-2017-decoupled}
Ilya Loshchilov and Frank Hutter. 2017.
\newblock Decoupled weight decay regularization.
\newblock \emph{arXiv preprint arXiv:1711.05101}.

\bibitem[{Miao et~al.(2020)Miao, Liang, and Su}]{miao-etal-2020-diverse}
Shen-yun Miao, Chao-Chun Liang, and Keh-Yih Su. 2020.
\newblock \href {https://doi.org/10.18653/v1/2020.acl-main.92} {A diverse corpus for evaluating and developing {E}nglish math word problem solvers}.
\newblock In \emph{Proceedings of the 58th Annual Meeting of the Association for Computational Linguistics}, pages 975--984.

\bibitem[{Mitra and Baral(2016)}]{mitra-baral-2016-learning}
Arindam Mitra and Chitta Baral. 2016.
\newblock \href {https://doi.org/10.18653/v1/P16-1202} {Learning to use formulas to solve simple arithmetic problems}.
\newblock In \emph{Proceedings of the 54th Annual Meeting of the Association for Computational Linguistics (Volume 1: Long Papers)}, pages 2144--2153.

\bibitem[{Patel et~al.(2021)Patel, Bhattamishra, and Goyal}]{patel-etal-2021-nlp}
Arkil Patel, Satwik Bhattamishra, and Navin Goyal. 2021.
\newblock \href {https://doi.org/10.18653/v1/2021.naacl-main.168} {Are {NLP} models really able to solve simple math word problems?}
\newblock In \emph{Proceedings of the 2021 Conference of the North American Chapter of the Association for Computational Linguistics: Human Language Technologies}, pages 2080--2094.

\bibitem[{Qin et~al.(2020)Qin, Lin, Liang, Zhang, and Lin}]{qin-etal-2020-semantically}
Jinghui Qin, Lihui Lin, Xiaodan Liang, Rumin Zhang, and Liang Lin. 2020.
\newblock \href {https://doi.org/10.18653/v1/2020.emnlp-main.309} {Semantically-aligned universal tree-structured solver for math word problems}.
\newblock In \emph{Proceedings of the 2020 Conference on Empirical Methods in Natural Language Processing (EMNLP)}, pages 3780--3789.

\bibitem[{Rittle-Johnson and Alibali(1999)}]{rittle1999conceptual}
Bethany Rittle-Johnson and Martha~Wagner Alibali. 1999.
\newblock Conceptual and procedural knowledge of mathematics: Does one lead to the other?
\newblock \emph{Journal of educational psychology}, 91(1):175.

\bibitem[{Rittle-Johnson and Schneider(2014)}]{rittle-etal-2014-developing}
Bethany Rittle-Johnson and Michael Schneider. 2014.
\newblock \href {https://doi.org/10.1093/oxfordhb/9780199642342.013.014} {Developing conceptual and procedural knowledge of mathematics}.
\newblock In \emph{The Oxford Handbook of Numerical Cognition}. Oxford University Press.

\bibitem[{Shen et~al.(2021)Shen, Yin, Li, Shang, Jiang, Zhang, and Liu}]{shen-etal-2021-generate-rank}
Jianhao Shen, Yichun Yin, Lin Li, Lifeng Shang, Xin Jiang, Ming Zhang, and Qun Liu. 2021.
\newblock \href {https://doi.org/10.18653/v1/2021.findings-emnlp.195} {Generate {\&} rank: A multi-task framework for math word problems}.
\newblock In \emph{Findings of the Association for Computational Linguistics: EMNLP 2021}, pages 2269--2279.

\bibitem[{Shen and Jin(2020)}]{shen-jin-2020-solving}
Yibin Shen and Cheqing Jin. 2020.
\newblock \href {https://doi.org/10.18653/v1/2020.coling-main.262} {Solving math word problems with multi-encoders and multi-decoders}.
\newblock In \emph{Proceedings of the 28th International Conference on Computational Linguistics}, pages 2924--2934.

\bibitem[{Vaswani et~al.(2017)Vaswani, Shazeer, Parmar, Uszkoreit, Jones, Gomez, Kaiser, and Polosukhin}]{vaswani-etal-2017-attention}
Ashish Vaswani, Noam Shazeer, Niki Parmar, Jakob Uszkoreit, Llion Jones, Aidan~N Gomez, \L~ukasz Kaiser, and Illia Polosukhin. 2017.
\newblock Attention is all you need.
\newblock In \emph{Advances in Neural Information Processing Systems}, volume~30.

\bibitem[{Wang et~al.(2018)Wang, Wang, Cai, Zhang, and Liu}]{wang-etal-2018-translating}
Lei Wang, Yan Wang, Deng Cai, Dongxiang Zhang, and Xiaojiang Liu. 2018.
\newblock \href {https://doi.org/10.18653/v1/D18-1132} {Translating a math word problem to a expression tree}.
\newblock In \emph{Proceedings of the 2018 Conference on Empirical Methods in Natural Language Processing}, pages 1064--1069.

\bibitem[{Wang et~al.(2017)Wang, Liu, and Shi}]{wang-etal-2017-deep}
Yan Wang, Xiaojiang Liu, and Shuming Shi. 2017.
\newblock \href {https://doi.org/10.18653/v1/D17-1088} {Deep neural solver for math word problems}.
\newblock In \emph{Proceedings of the 2017 Conference on Empirical Methods in Natural Language Processing}, pages 845--854.

\bibitem[{Wei et~al.(2022)Wei, Wang, Schuurmans, Bosma, brian ichter, Xia, Chi, Le, and Zhou}]{wei-etal-2022-chain}
Jason Wei, Xuezhi Wang, Dale Schuurmans, Maarten Bosma, brian ichter, Fei Xia, Ed~H. Chi, Quoc~V Le, and Denny Zhou. 2022.
\newblock Chain of thought prompting elicits reasoning in large language models.
\newblock In \emph{Advances in Neural Information Processing Systems}.

\bibitem[{Wu et~al.(2021)Wu, Zhang, and Wei}]{wu-etal-2021-edge-enhanced}
Qinzhuo Wu, Qi~Zhang, and Zhongyu Wei. 2021.
\newblock \href {https://doi.org/10.18653/v1/2021.findings-emnlp.127} {An edge-enhanced hierarchical graph-to-tree network for math word problem solving}.
\newblock In \emph{Findings of the Association for Computational Linguistics: EMNLP 2021}, pages 1473--1482.

\bibitem[{Xie and Sun(2019)}]{xie-sun-2019-goal}
Zhipeng Xie and Shichao Sun. 2019.
\newblock \href {https://doi.org/10.24963/ijcai.2019/736} {A goal-driven tree-structured neural model for math word problems}.
\newblock In \emph{Proceedings of the Twenty-Eighth International Joint Conference on Artificial Intelligence, {IJCAI-19}}, pages 5299--5305.

\bibitem[{Xiong et~al.(2020)Xiong, Yang, He, Zheng, Zheng, Xing, Zhang, Lan, Wang, and Liu}]{xiong-etal-2020-layer}
Ruibin Xiong, Yunchang Yang, Di~He, Kai Zheng, Shuxin Zheng, Chen Xing, Huishuai Zhang, Yanyan Lan, Liwei Wang, and Tieyan Liu. 2020.
\newblock \href {https://proceedings.mlr.press/v119/xiong20b.html} {On layer normalization in the transformer architecture}.
\newblock In \emph{Proceedings of the 37th International Conference on Machine Learning}, volume 119, pages 10524--10533.

\bibitem[{Yang et~al.(2022)Yang, Qin, Chen, and Liang}]{yang-etal-2022-unbiased}
Zhicheng Yang, Jinghui Qin, Jiaqi Chen, and Xiaodan Liang. 2022.
\newblock \href {https://doi.org/10.18653/v1/2022.findings-naacl.104} {Unbiased math word problems benchmark for mitigating solving bias}.
\newblock In \emph{Findings of the Association for Computational Linguistics: NAACL 2022}, pages 1401--1408.

\bibitem[{Yu et~al.(2021)Yu, Wen, Zheng, and Xiao}]{yu-etal-2021-improving}
Weijiang Yu, Yingpeng Wen, Fudan Zheng, and Nong Xiao. 2021.
\newblock \href {https://doi.org/10.18653/v1/2021.emnlp-main.272} {Improving math word problems with pre-trained knowledge and hierarchical reasoning}.
\newblock In \emph{Proceedings of the 2021 Conference on Empirical Methods in Natural Language Processing}, pages 3384--3394.

\bibitem[{Zhang et~al.(2020{\natexlab{a}})Zhang, Lee, Lim, Qin, Wang, Shao, and Sun}]{zhang-etal-2020-teacher}
Jipeng Zhang, Roy Ka-Wei Lee, Ee-Peng Lim, Wei Qin, Lei Wang, Jie Shao, and Qianru Sun. 2020{\natexlab{a}}.
\newblock Teacher-student networks with multiple decoders for solving math word problem.
\newblock In \emph{Proceedings of the Twenty-Ninth International Joint Conference on Artificial Intelligence, {IJCAI-20}}, pages 4011--4017.

\bibitem[{Zhang et~al.(2020{\natexlab{b}})Zhang, Wang, Lee, Bin, Wang, Shao, and Lim}]{zhang-etal-2020-graph-tree}
Jipeng Zhang, Lei Wang, Roy Ka-Wei Lee, Yi~Bin, Yan Wang, Jie Shao, and Ee-Peng Lim. 2020{\natexlab{b}}.
\newblock \href {https://doi.org/10.18653/v1/2020.acl-main.362} {Graph-to-tree learning for solving math word problems}.
\newblock In \emph{Proceedings of the 58th Annual Meeting of the Association for Computational Linguistics}, pages 3928--3937.

\end{thebibliography}
